\documentclass{acm_proc_article-sp}

\usepackage{enumerate}

\begin{document}


\title{Scalability of Genetic Programming and Probabilistic Incremental Program Evolution}



\numberofauthors{3}
\author{
\alignauthor Radovan Ondas\\
       \affaddr{University of Missouri - St. Louis}\\
       \affaddr{Dept. of Math and Computer Science, CCB 331}\\
       \affaddr{8001 Natural Bridge Rd.}\\
       \affaddr{St. Louis, MO 63121, USA}\\
       \email{ondasr@umsl.edu}
\alignauthor Martin Pelikan\\
       \affaddr{University of Missouri - St. Louis}\\
       \affaddr{Dept. of Math and Computer Science CCB 320}\\
       \affaddr{8001 Natural Bridge Rd.}\\
       \affaddr{St. Louis, MO 63121, USA}\\
       \email{pelikan@cs.umsl.edu}
\alignauthor Kumara Sastry\\
       \affaddr{University of Illinois at Urbana-Champaign}\\
       \affaddr{Dept. of General Engineering}\\
       \affaddr{117 Transportation Bldg.}\\
       \affaddr{104 S. Mathews Ave.}\\
       \affaddr{Urbana, IL 61801, USA}\\
       \email{ksastry@uiuc.edu}       
}

\date{10 January 2005}
\maketitle

\begin{abstract}
This paper discusses scalability of standard genetic programming (GP) and the probabilistic incremental program evolution (PIPE). To investigate the need for both effective mixing and linkage learning, two test problems are considered: \texttt{ORDER} problem, which is rather easy for any recombination-based GP, and \texttt{TRAP} or the deceptive trap problem, which requires the algorithm to learn interactions among subsets of terminals. The scalability results show that both GP and PIPE scale up polynomially with problem size on the simple \texttt{ORDER} problem, but they both scale up exponentially on the deceptive problem. This indicates that while standard recombination is sufficient when no interactions need to be considered, for some problems linkage learning is necessary. These results are in agreement with the lessons learned in the domain of binary-string genetic algorithms (GAs). Furthermore, the paper investigates the effects of introducing unnecessary and irrelevant primitives on the performance of GP and PIPE.
\end{abstract}

\category{\\I.2.6}{Artificial Intelligence}{Learning}
\category{\\G.1.6}{Numerical Analysis}{Optimization}



\section{Introduction}
To solve large and complex problems, scalability is among the primary concerns of an optimization practitioner. However, only few studies~\cite{sastry:pmbcgp,Sastry:04} exist that study scalability in genetic programming (GP)\cite{koza:gp}. The same holds for simple approaches to using probabilistic recombination in GP within the estimation of distribution algorithm (EDA) framework~\cite{Muhlenbein:96**,Larranaga:02,Pelikan:02}, such as the probabilistic incremental program evolution (PIPE)~\cite{Salustowicz:98}.

The purpose of this paper is to study the scalability of standard GP and PIPE on two decomposable GP problems: \texttt{ORDER} and \texttt{TRAP}. The two algorithms perform as expected and they solve~\texttt{ORDER} scalably while failing to scale up on \texttt{TRAP}. Additionally, the paper studies the effects of introducing unnecessary and irrelevant primitives. Both GP and PIPE are shown to deal with these two sources of difficulty well. The results presented in this paper confirm that binary-string GAs have a lot in common with GP and PIPE, and thus the lessons learned in the design, study, and application of standard GAs and their extensions should carry over to GP as argued for example in~\cite{Goldberg:98,sastry:pmbcgp,Sastry:04}.

The paper starts by describing the algorithms investigated in this paper: GP and PIPE. Section~\ref{section-test-problems} explains test problems. Section~\ref{section-experiments} provides and discusses experimental results. Section~\ref{section-future-work} presents important topics for future work in this line of research. Section~\ref{section-summary} summarizes the paper. Finally, Section~\ref{section-conclusions} concludes the paper.

\section{Methods}
\begin{figure*}
\centering
\epsfig{file=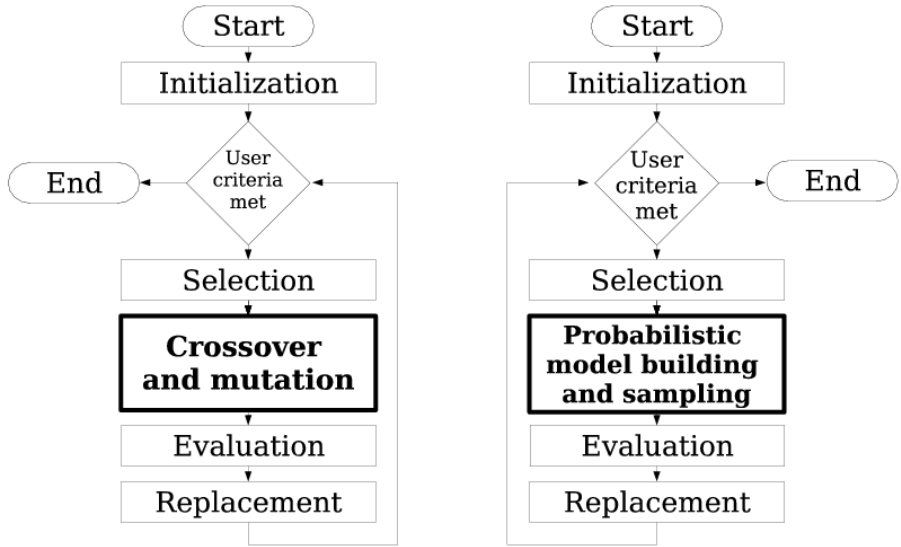, width=12cm}
\caption{Standard genetic programming (GP) and the probabilistic incremental program evolution (PIPE).}
\label{fig-gp-pipe}
\end{figure*}

Both GP and PIPE work with programs encoded as labeled-tree structures and both can be applied to the same class of problems. While GP generates new candidate programs using standard variation operators, such as crossover and mutation, PIPE builds and samples a probabilistic model in the form of a tree of mutually independent nodes. Therefore, the difference between GP and PIPE is in their variation operator (see Figure~\ref{fig-gp-pipe}).

This section describes GP and PIPE. 
The section starts by discussing standard GP and closes by describing
the probabilistic algorithm PIPE.

\subsection{Genetic Programming}
Genetic programming (GP) \cite{koza:gp} is a genetic algorithm (GA) \cite{Goldberg:89d} that evolves programs instead of fixed-length strings. Programs are represented by trees where nodes represent functions and leaves represent variables and constants. 

GP starts with a population of random candidate programs. Each program is evaluated on a given task and its fitness value is assigned. A population of promising programs is then selected using one of the standard GA selection operators, such as tournament or truncation selection. Some of the selected programs can be directly copied into the new population, the remaining ones are copied after applying variation operators, such as crossover and mutation. Crossover usually proceeds by exchanging randomly selected subtrees between two programs, whereas mutation usually replaces a randomly selected subtree of a program by a randomly generated one. This process is repeated until termination criteria are met.

Since standard GP variation operators proceed without considering interactions between different components of selected programs, they are likely to experience difficulties with solving problems where different program components interact strongly. However, problems that can be decomposed into subproblems of order one should be easy for any standard GP based on recombination. This intuition is verified with experiments in Section~\ref{section-experiments}. Similar behavior can be observed in GAs; GAs with standard variation operators work great on problems with no interactions between decision variables~\cite{Muhlenbein:93c, Harik:97a**, Goldberg:book}, but they often fail for problems with highly interacting decision variables~\cite{Thierens:95, Goldberg:book}.

We implemented GP using the \texttt{lilgp} GP library developed by the Genetic Algorithms Research and Applications Group (GARAGe) at the Michigan State University.

\subsection{PIPE}
In the probabilistic incremental program evolution (PIPE) algorithm \cite{salust:pipe, Salustowicz:98} computer programs or mathematical expressions are evolved like in GP~\cite{koza:gp}. However, pairwise crossover and mutation are replaced by building a probabilistic model of promising programs and sampling the model. 

Like GP, PIPE represents programs by labeled trees where each internal node represents a function and each leaf represents a variable or a constant. The initial population is also generated at random. All programs in the population are then evaluated and selection is applied to select the population of promising programs. Instead of applying crossover and mutation to a part of the selected population to generate new programs, PIPE now builds a probabilistic model of the selected programs in the form of a tree. This probabilistic model is then sampled to generate new candidate programs that form the new population. The process is repeated until the termination criteria are met.

Next, the methods for learning and sampling the probabilistic model in PIPE are described. 

\subsubsection{Learning the Probabilistic Model}
\begin{figure}
\centering
\epsfig{file=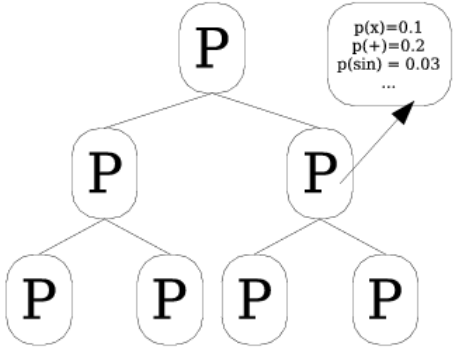, width=0.35\textwidth}
\caption{A probabilistic model of a population of programs in the form of a tree with nodes representing the probabilities of functions and terminals. All nodes are modeled independently.}
\label{fig-pipe-model}
\end{figure}

The probabilistic model in PIPE is a tree with the structure corresponding to the structure of candidate programs. Since different programs may be of different structure and size, the population is first parsed to find the smallest tree that contains every structure in the selected population. Each node of a program in the selected population then directly corresponds to one node in the model, whereas the children of each internal node represent arguments of the function in this node. Figure~\ref{fig-pipe-model} illustrates probabilistic models used in PIPE.

If there are functions of different arities, the number of children of each node in the probabilistic model is equal to the maximum arity of a function in this node in the selected population. For a function of smaller arity, the first children are interpreted as arguments of this function (in an arbitrary fixed ordering). 

PIPE then parses the selected population and computes the probabilities of different functions and terminals in each node of the probabilistic model. The nodes of the probabilistic model thus consist of tables of probabilities, and there is one probability for each function or terminal in each node. 

\subsubsection{Sampling the Probabilistic Model}

Sampling of the probabilistic model starts in the root of the probabilistic model. The same recursive procedure is used to generate each node. First, a function or terminal is generated in the current node based on the distribution encoded by the table of probabilities in this node. If the function requires several arguments, a necessary number of children are generated recursively. The recursive generation terminates in a node whenever a terminal is generated in this node and thus no children have to be generated. Since the probabilistic model is built from an actual population of programs, the sampling will never cross the boundaries of the model. 
 
Using the probabilistic model of PIPE to model and sample candidate programs resembles the univariate marginal distribution algorithm (UMDA)~\cite{Muhlenbein:96**, Baluja:94}, which models each string position independently of the values in other positions. Interactions between each node and its context are ignored. That is why it can be expected that using this model will lead to inferior results on problems where program components interact strongly, similarly as the univariate model generally fails if string positions interact~\cite{Thierens:95}. On the other hand, if different program components are mutually independent, PIPE should work great. This intuition is verified with experiments in Section~\ref{section-experiments}. 

We implemented PIPE by incorporating probabilistic recombination into the \texttt{lilgp} library developed by GARAGe at the Michigan State University.

\section{Test problems}
\label{section-test-problems}
In order to test scalability, we need a class of problems where size can be modified while the inherent problem difficulty does not grow prohibitively fast. In fixed-length string GAs, decomposable problems of bounded difficulty~\cite{Goldberg:book} can be used as a challenging but solvable class of problems. Two types of decomposable problems for fixed-length string GAs are common: Onemax and concatenated traps. In onemax, the contribution of each bit is independent of its context. On the other hand, in concatenated traps, bits in each trap partition interact and cannot be effectively processed without considering other bits in the same trap partition. 

Similar problems to onemax and concatenated traps were also created for GP where candidate solutions are represented by program trees~\cite{Goldberg:98,sastry:pmbcgp}. Two classes of problems from \cite{sastry:pmbcgp} are considered:
\begin{enumerate}
   \item \texttt{ORDER}: OneMax-like, GP-easy problem.
   \item \texttt{TRAP}: Deceptive-trap-like, GP-difficult problem
\end{enumerate}
\texttt{ORDER} should be easy for any recombination-based GP. However, since standard variation operators do not consider interactions between different program components, \texttt{TRAP} can be expected to lead to exponential scalability of both standard GP and PIPE. The problems are described next.

\subsection{Problem 1: Order}
The primitive set of an $l$-primitive \texttt{ORDER} problem consist of a binary function \texttt{JOIN} and complimentary terminals $X_i$ and $\overline{X}_i$ for $i\in\{1, 2, \ldots, l\}$. A candidate solution of the \texttt{ORDER} problem is a binary tree  with \texttt{JOIN} in all internal nodes and either $X_i$'s or $\overline{X}_i$'s at its leaves. The candidate solution's output is determined  by parsing  the program tree inorder (from left to right). The program expresses $X_i$ if, during the inorder  parse, $X_i$ is encountered before its complement $\overline{X}_i$ and neither $X_i$ nor its complement are encountered earlier. For all $i\in\{1, 2, \ldots, l\}$, if $X_i$ is unexpressed, $\overline{X}_i$ is expressed instead. One terminal is thus expressed from each pair $X_i$ and $\overline{X}_i$.

For all $i\in\{1, 2, \ldots, l\}$, an equal unit of fitness value is accredited if $X_i$ is expressed:
\begin{equation}
   f_1(x_i) = \{ \begin{array}{ll}
      1 & \mbox{if}~x_i \in \{X_1, X_2, \dots, X_l\}\\
      0 & \mbox{otherwise}
   \end{array}
\end{equation}

The fitness function for \texttt{ORDER} is defined as
\begin{equation}
   F(x) = \sum_{i=1}^l f_1(x_i), 
\end{equation}
where $x$ is the set of primitives expressed by the program. Given that trees can be sufficiently large, the expression for a globally optimal solution of an $l-$primitive \texttt{ORDER} problem is $\{X_1, X_2, \dots, X_l\}$ and thus its fitness value is $l$.

\begin{figure}
\centering
\epsfig{file=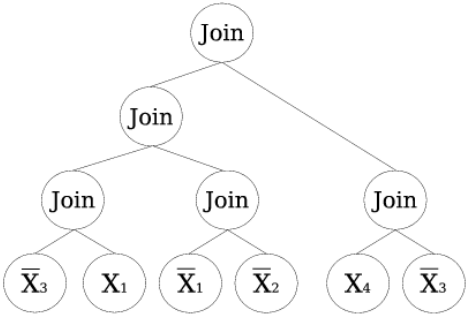, width=0.35\textwidth}
\caption{A candidate solution for a 4-primitive \texttt{ORDER} problem. The output of the program is $\{X_1, \overline{X}_2, \overline{X}_3, X_4\}$ and the fitness of this solution is thus $2$.}
\label{fig-order-example}
\end{figure}

For example, consider a candidate solution for a 4-primitive \texttt{ORDER} problem shown in Figure~\ref{fig-order-example}. The sequence of leaves visited during the inorder parse is $\{\overline{X}_3, X_1, \overline{X}_1, \overline{X}_2, X_4, \overline{X}_3\}$, the expression of this sequence is $\{X_1, \overline{X}_2, \overline{X}_3, X_4\}$, and the fitness of this solution is thus $2$.

\subsection{Problem 2: Deceptive Trap}
In standard GAs, deceptive functions~\cite{deb:intro, Goldberg:book} are designed to thwart the very mechanism of selectorecombinative search by punishing any localized hillclimbing and requiring mixing of whole building blocks at or above the order of deception. Using such adversarially designed functions is a stiff test---in some sense the stiffest test---of algorithm performance. The idea is that if an algorithm can beat an adversarially designed test functions, it can solve other problems that are equally hard  or easier than the adversary. Furthermore, if the building blocks of such deceptive functions are not identified and respected by selectorecombinative GAs, then they almost always converge to the local minimum.

\texttt{TRAP} is designed to test the same mechanisms in GP. Fitness is computed so that if interactions between different components of the program are not considered, optimization may be mislead away from the global optimum. Similarly as with standard GAs on deceptive functions, standard GP is expected to fail in solving \texttt{TRAP} scalably, indicating the need for linkage learning in GP.

Programs in \texttt{TRAP} also consist of one binary function \texttt{JOIN} and $l$ pairs of complementary primitives $X_i$ and $\overline{X}_i$. The expression mechanism  of the program for \texttt{TRAP} is identical to that to that of \texttt{ORDER}. The difference is in the fitness evaluation procedure. 

In \texttt{TRAP}, the expressed set of primitives is first mapped to an $l$-bit binary string. The $i$th bit of the string is $1$ if and only if $X_i$ was expressed; otherwise, the $i$th bit of the string is $0$. The resulting binary string is then partitioned into groups of $k$ bits each (the partitioning is fixed during the entire run) and a trap function is applied to each group:
\begin{equation}
   f_k(u) = \{ 
   \begin{array}{ll}
      1.0 & u = k\\
      (1.0 - \delta)(1 - \frac{u}{k-1}) & u < k 
   \end{array}
\end{equation}
where $u$ is the number of ones in the input string of $k$ bits.

\begin{figure}
\centering
\epsfig{file=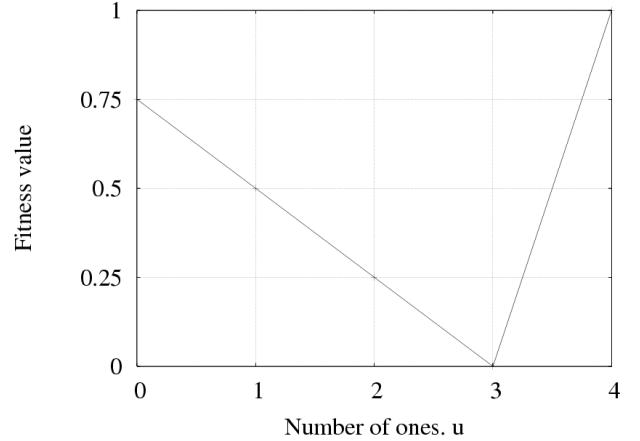, width=0.49\textwidth}
\caption{A fully deceptive trap function with $k = 4$, and $\delta = 0.25$.}
\label{fig-trap}
\end{figure}

The fitness function of the trap function is then computed by adding the contributions of all groups of $k$ bits together. 

The difficulty of \texttt{trap} can be adjusted by modifying the values $k$, and $\delta$. The problem becomes more difficult  as the value of $k$ is increased and that of $\delta$ is decreased. A $4$-bit deceptive trap function is illustrated in Figure~\ref{fig-trap}. In this paper we use traps with $k=3$ and $\delta=1$. 

The important feature of additively separable trap functions is that if looking at the performance of any subset of $k$ bits corresponding to one trap, it seems to be better to propagate $0$s (here we need to eliminate $X_i$s and substitute $\overline{X}_i$ or nothing). As shown in~\cite{sastry:pmbcgp}, if interactions between different components of the program are not considered, it can be expected that GP will scale up poorly on this problem.

\subsection{Other primitives}
In addition to \texttt{ORDER} and \texttt{TRAP} with \texttt{JOIN} and $l$ terminal pairs, we tested GP and PIPE on \texttt{ORDER} with additional two primitives: A primitive negative join and junk or unexpressed terminals. The purpose of additional tests was to determine how GP and PIPE respond to more complex interactions and unnecessary program primitives. 

\subsubsection{Primitive negative-join}
\texttt{NEG\_JOIN} affects all its descendant terminals by expressing each primitive $X_i$ as its negation $\overline{X}_i$; analogically, all descendants $\overline{X}_i$ are expressed as $X_i$. If a terminal has more \texttt{NEG\_JOIN} ancestors, only one of them is considered and the terminal is negated only once.

\texttt{NEG\_JOIN} is unnecessary for solving \texttt{ORDER} and it does not introduce a less complex or easier to find global optimum. Furthermore, \texttt{NEG\_JOIN} introduces interactions into \texttt{ORDER} because the best value in each leaf depends on its ancestors. Nonetheless, these interactions are relatively simple as many leaves are expected to contain \texttt{NEG\_JOIN} on the path to the root.

\begin{figure}
\centering
\epsfig{file=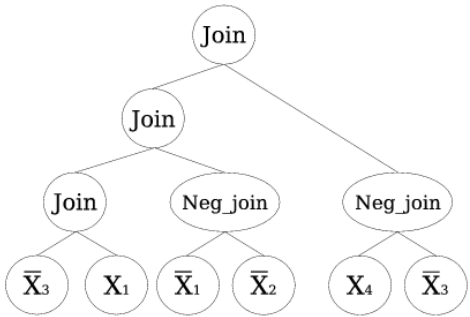, width=0.35\textwidth}
\caption{A candidate solution for 4-primitive \texttt{ORDER} problem with \texttt{NEG\_JOIN}. The output of the program is $\{X_1, X_2, X_3, \overline{X}_4\}$, and the fitness of this solution is thus $3$.}
\label{fig-neg-example}
\end{figure}

For example, for the program shown in Figure~\ref{fig-neg-example}, the inorder pass through the program results in the following sequence of leaves: $\{\overline{X}_3, X_1, X_1, X_2, \overline{X}_4, X_3\}$. The expression gives us $\{X_1, X_2, X_3, \overline{X}_4\}$, and thus the fitness is $3$.

\subsubsection{Junk-code terminals}
Junk-code or \texttt{JUNK} terminals represent unnecessary primitives that are irrelevant for the particular problem. In biological terms, \texttt{JUNK} terminals correspond to junk code in DNA. During the expression phase, \texttt{JUNK} terminals are simply ignored and they thus do not influence the overall fitness at all. 

Adding \texttt{JUNK} terminals makes the optimization problem more difficult, because additional primitives enlarge the search space without simplifying the problem. The influence of \texttt{JUNK} terminals can be tuned by changing the number of unique \texttt{JUNK} terminals.

\begin{figure}
\centering
\epsfig{file=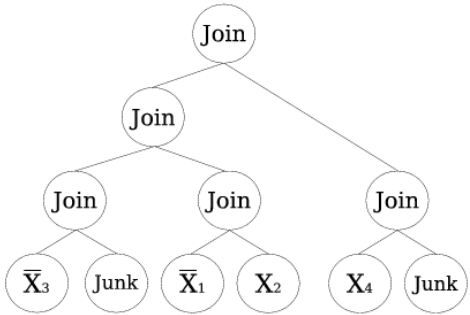, width=0.35\textwidth}
\caption{A candidate solution for 4-primitive \texttt{ORDER} problem with \texttt{JUNK} terminals. The output of the program is $\{\overline{X}_1, X_2, \overline{X}_3, X_4\}$, and the fitness of this solution is thus $2$.}
\label{fig-junk-example}
\end{figure}

Figure~\ref{fig-junk-example} shows a tree with two \texttt{JUNK} terminals. The inorder parse results in the following sequence of leaves (ignoring \texttt{JUNK}): $\{\overline{X}_3, \overline{X}_1, X_2, X_4\}$. The expression gives us $\{\overline{X}_1, X_2, \overline{X}_3, X_4\}$, and thus the fitness of this solution is $2$. 

\section{Experiments}
\label{section-experiments}
This section compares the performance of GP and PIPE on three variants of \texttt{ORDER} and one variant of \texttt{TRAP}. 

\subsection{Description of experiments}
The scalability of GP and PIPE was tested on four classes of problems:
\begin{enumerate}[(i)]
\item Basic \texttt{ORDER} (no \texttt{JUNK} or \texttt{NEG\_JOIN}),
\item basic \texttt{TRAP} (no \texttt{JUNK} or \texttt{NEG\_JOIN}),
\item \texttt{ORDER} with \texttt{NEG\_JOIN}, and
\item \texttt{ORDER} with \texttt{JUNK} terminals, where the number of unique \texttt{JUNK} terminals is set to $l/5$. 
\end{enumerate}
The scalability experiments were performed by testing both algorithms on problem instances with an increasing number of primitives.

Additionally, the effects of increasing the number of unnecessary primitives on the performance of GP and PIPE were studied by testing GP and PIPE on a $20$-primitive \texttt{ORDER} with an increasing number of \texttt{JUNK} terminals (from $5$ to $40$). 

Binary tournament selection was used in both GP and PIPE. The probability of crossover in GP is set to $1.0$. To focus on the effects of recombination, no mutation is used. The initial population in both methods was generated using the standard half-and-half method. Maximum tree depth was set to be one more than the depth of the minimum tree to store the global optimum. The population size that is within $10\%$ of the minimum population size required to solve 30 independent runs is used. The population size is determined using a bisection method. The runs are terminated when the algorithms find the global optimum or when the number of generations is too large for the particular problem.

\subsection{Results}
\begin{figure}
\centering
\epsfig{file=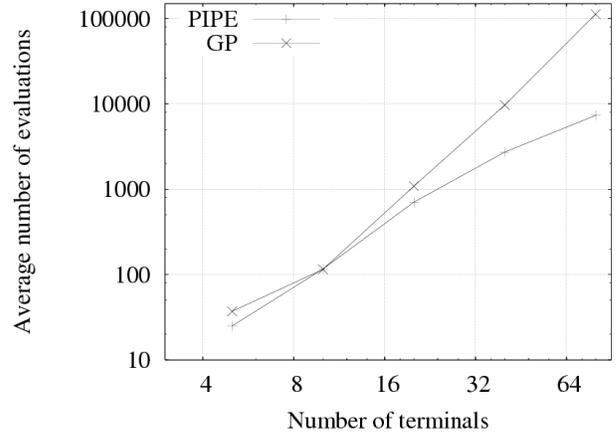, width=0.49\textwidth}
\caption{Scalability of GP and PIPE on \texttt{ORDER}.}
\label{fig-order-exp}
\end{figure}

Figure~\ref{fig-order-exp} shows the scalability of GP and PIPE on \texttt{ORDER} without \texttt{NEG\_JOIN} or \texttt{JUNK} terminals. Problem instances of different size were examined; more specifically, $l=5$, $10$, $20$, $40$, $60$, $80$, and $100$. The figure shows the average number of function evaluations of 30 successful runs with respect to the problem size (number of positive literals). The results indicate that PIPE is slightly more efficient than GP but both GP and PIPE scale up with a low-order polynomial. These results are in agreement with the behavior observed in binary-string GAs on the simple onemax problem. On onemax, both simple GA and UMDA find the optimum in low-order polynomial time~\cite{Muhlenbein:93c,Harik:97a**,Goldberg:book,Pelikan:02a}; however, UMDA performs slightly better~\cite{Pelikan:02a} because it uses a more effective recombination for this type of problems. 

\begin{figure}
\centering
\epsfig{file=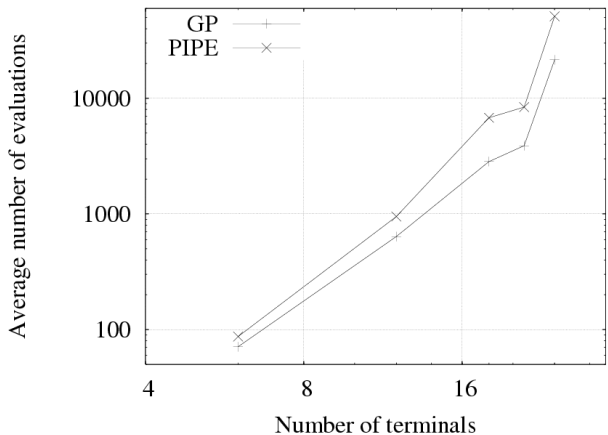, width=0.49\textwidth}
\caption{Scalability of GP and PIPE on \texttt{TRAP}.}
\label{fig-trap-exp}
\end{figure}

Figure~\ref{fig-trap-exp} compares the scalability of GP and the PIPE on \texttt{TRAP} without \texttt{NEG\_JOIN} or \texttt{JUNK} terminals. The size of one trap is $k=3$ and the signal difference is $d=1$. Problem instances of different size were examined; more specifically, $l=6$, $12$, $18$, $21$, $24$, and $33$. On \texttt{TRAP}, GP performs slightly better than PIPE. This can be explained by its weaker recombination operator because here recombination causes disruption of important partial solutions~\cite{Thierens:95} as can be hypothesized based on the performance of standard GAs on similar problems. Nonetheless, both GP and PIPE scale up poorly and they indicate an exponential growth of the number of function evaluations with problem size. 

\begin{figure}
\centering
\epsfig{file=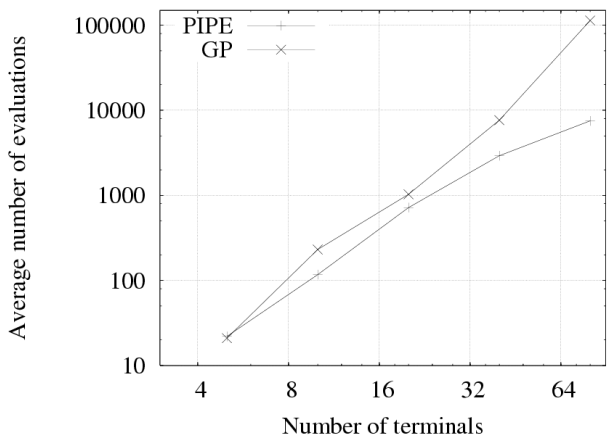, width=0.49\textwidth}
\caption{Scalability of GP and PIPE on \texttt{ORDER} with \texttt{NEG\_JOIN}.}
\label{fig-neg-join-exp}
\end{figure}

Figure~\ref{fig-neg-join-exp} compares the scalability of GP and PIPE on \texttt{ORDER} with \texttt{NEG\_JOIN}. Problem instances of different size were examined; more specifically, $l=5$, $10$, $20$, $40$, $60$, $80$, and $100$. Both GP and PIPE perform similarly as on basic \texttt{ORDER} without \texttt{NEG\_JOIN}, but there is a slight decrease in their performance because of the interactions introduced by \texttt{NEG\_JOIN}.

\begin{figure}
\centering
\epsfig{file=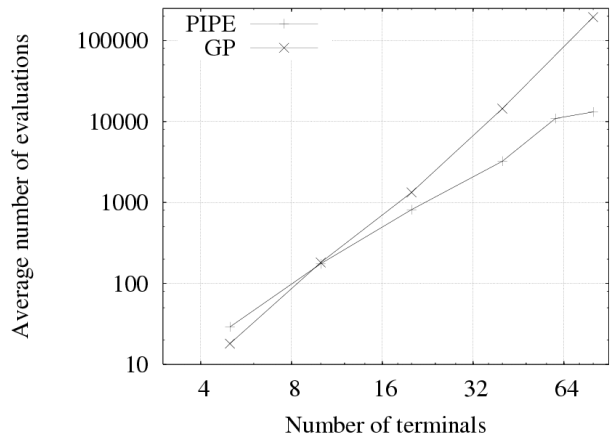, width=0.49\textwidth}
\caption{Scalability of GP and PIPE on \texttt{ORDER} with $l/5$ copies of \texttt{JUNK} terminals.}
\label{fig-junk-exp}
\end{figure}

Figure~\ref{fig-junk-exp} compares the scalability of GP and PIPE on \texttt{ORDER} with $l/5$ unique \texttt{JUNK} terminals. For example, a problem instance with $l=20$ positive terminals contains $4$ unique \texttt{JUNK} terminals. Both GP and PIPE seem to be capable of dealing with these irrelevant terminals and achieve performance comparable to that on basic \texttt{ORDER}. 

\begin{figure}
\centering
\epsfig{file=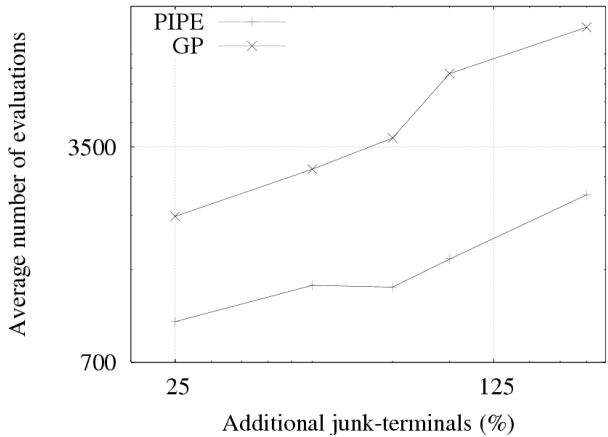, width=0.49\textwidth}
\caption{Scalability of GP and PIPE on \texttt{ORDER} with an increasing number of \texttt{JUNK} terminals (from $5$ to $40$). The maximum depth of candidate programs is $6$.}
\label{fig-junk64-exp}
\end{figure}

\begin{figure}
\centering
\epsfig{file=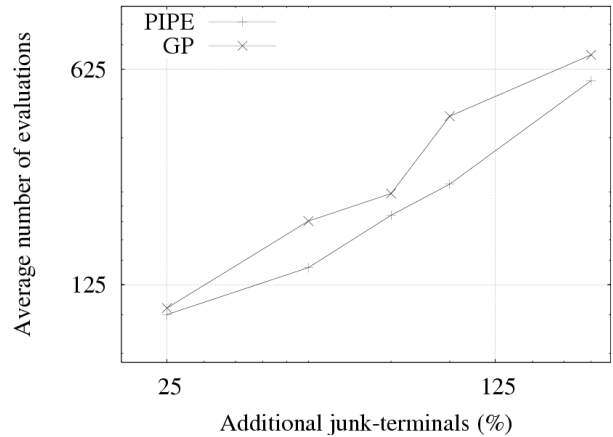, width=0.49\textwidth}
\caption{Scalability of GP and PIPE on \texttt{ORDER} with an increasing number of \texttt{JUNK} terminals (from $5$ to $40$). The maximum depth of candidate programs is $7$.}
\label{fig-junk128-exp}
\end{figure}

The last two sets of experiments are similar in that they show how the performance of GP and PIPE changes when adding irrelevant terminals into the representation. \texttt{ORDER} with $l=20$ terminals is used with the number of \texttt{JUNK} terminals ranging from $5$ to $40$ ($5$, $10$, $15$, $20$, and $40$). The experiments differ in the bound on the maximum tree depth. Figure~\ref{fig-junk64-exp} shows the results with the depth limited to at most $6$ (so there are at most $7$ levels including the root). Figure~\ref{fig-junk128-exp} shows the results with the depth limited to at most $7$ (so there are at most $8$ levels including the root). The problem with the smaller maximum depth is more difficult for both GP and PIPE because \texttt{JUNK} terminals obstruct the creation of an optimal solution that is only slightly larger than the maximum allowed tree. PIPE deals better with this ``lack of space'' than GP does. However, in both cases, the number of evaluations still appears to grow with a low-order polynomial or slower as irrelevant terminals are added. 

\section{Future Work}
\label{section-future-work}
Future work should study the scalability of GP, PIPE, and other similar approaches on the problems presented in this paper and other problems where problem size can be modified without affecting the inherent problem difficulty. The efforts to introducing linkage learning into GP (for example ~\cite{sastry:pmbcgp,Bosman:04}) should continue to succeed in the design of robust GP methods that provide a scalable solution to broad classes of GP problems. Finally, more theory should be designed to match the achievements in this area in the domain of GAs~\cite{Muhlenbein:93c,Thierens:95,Goldberg:book,Pelikan:02a,MillerBL:96,Lobo:00}.

\section{Summary}
\label{section-summary}
This paper focused on the scalability of two GP algorithms: standard GP and PIPE. 

Two basic test functions were used: \texttt{ORDER} and \texttt{TRAP}. Both functions were defined using one binary function~\texttt{JOIN} and $l$ complementary terminal pairs $X_i$ and $\overline{X}_i$ for $i\in\{1, 2, \ldots, l\}$. \texttt{ORDER} can be solved without considering interactions between different program components, whereas \texttt{TRAP} introduces strong interactions, which make this function difficult for both standard crossover and mutation of GP, as well as the probabilistic recombination of PIPE.

The scalability of GP and PIPE was tested on basic \texttt{ORDER} and \texttt{TRAP}. Additionally, \texttt{ORDER} was extended by adding either of the following two primitives: (1) a binary function \texttt{NEG\_JOIN} and (2) \texttt{JUNK} (or irrelevant) terminals. Thus, there were 4 problem types examined.
 
On all four problem types, the scalability of GP and PIPE was first tested by applying these algorithms to problem instances of different size (number $l$ of positive terminals). Then, the sensitivity of GP and PIPE to the proportion of irrelevant terminals to the relevant ones was examined.

\section{Conclusions}
\label{section-conclusions}
The results presented in this paper indicate that the behavior of different variants of GP can be expected to be similar to that of standard binary-string GAs. There are two important consequences of this fact. First, as it was indicated in~\cite{sastry:pmbcgp}, to solve some classes of problems scalably, linkage learning may have to be incorporated into GP in order to identify and exploit interactions between different program components. Second, the lessons learned in the design and application of binary-string GAs should carry over to GP as argued for example in~\cite{Goldberg:98,Sastry:04}; the first steps along this direction are represented by the decision-making model of the population sizing in GP~\cite{Sastry:04}, which was based on the decision-making population-sizing model for standard GAs~\cite{Harik:97a**,Goldberg:book}.

The results also indicate that if the recombination operator captures interactions in the problem properly, increasing the mixing effects of recombination leads to better performance. That is why PIPE outperformed standard GP on problems where program components could be treated independently. This fact together with the need for linkage learning should encourage the application of probabilistic recombination operators of estimation of distribution algorithms (EDAs)~\cite{Muhlenbein:96**,Larranaga:02,Pelikan:02} to the domain of GP. Some representatives of EDAs applied to the GP domain are~\cite{salust:pipe,Salustowicz:98,sastry:pmbcgp,Bosman:04}.

Finally, the results show that both GP and PIPE can deal with irrelevant terminals and unnecessary functions relatively well and their performance gets only slightly worse when adding these primitives.

\section*{Acknowledgments}
This work was partially supported by the Research Award and the Research Board at the University of Missouri. Some experiments were done using the hBOA software developed by Martin Pelikan and David E. Goldberg at the University of Illinois at Urbana-Champaign. 
This work was also sponsored by the Air Force Office of Scientific Research, Air Force Materiel Command, USAF, under grant F49620-03-1-0129, the National Science Foundation under ITR grant DMR-99-76550 (at Materials Computation Center), and ITR grant DMR-0121695 (at CPSD), and the Dept. of Energy under grant DEFG02-91ER45439 (at Fredrick Seitz MRL). The U.S.  Government is authorized to reproduce and distribute reprints for government purposes notwithstanding any copyright notation thereon.


\bibliographystyle{abbrv}
\bibliography{bibliogr}  
%

\end{document}